\documentclass{article}

\usepackage{iftex}
\ifXeTeX
  \usepackage{xeCJK}
\fi

\usepackage{lmodern}
\usepackage{textalpha}     % For \textalpha, \textbeta etc.

\PassOptionsToPackage{numbers, compress}{natbib}
\usepackage[preprint]{neurips_2025}
\usepackage{natbib} % Explicitly load natbib

\usepackage[utf8]{inputenc} % allow utf-8 input
\usepackage[T1]{fontenc}    % use 8-bit T1 fonts

\usepackage{subcaption}

\usepackage{CJKutf8}      % Preferred CJK package for pdfLaTeX over the bare 'CJK'
\usepackage{siunitx} % For \num and related commands
\usepackage{textgreek}    % support for Greek letters in text
\usepackage{hyperref}     % hyperlinks
\usepackage{url}          % simple URL typesetting
\usepackage{booktabs}     % professional-quality tables
\usepackage{amsfonts}     % blackboard math symbols
\usepackage{longtable}    % long tables with booktabs support
\usepackage{wrapfig}      % wrapfigure and wraptable environments
\usepackage{subcaption}   % for subfigure environment
\usepackage{multirow}     % multi-row cells in tables
\usepackage{amsmath}      % for \text and other math features
\usepackage{amssymb}
\usepackage{nicefrac}     % compact symbols for 1/2, etc.
\usepackage{microtype}    % microtypography
\usepackage{xcolor}       % colors
\usepackage{graphicx}
\usepackage{enumitem}
\usepackage{float}
\usepackage{placeins}

\setlist[itemize]{nosep, left=0pt, labelsep=0.5em, itemsep=0pt, topsep=0pt, parsep=0pt, partopsep=0pt}
\setlist[enumerate]{nosep, left=0pt, labelsep=0.5em, itemsep=0pt, topsep=0pt, parsep=0pt, partopsep=0pt}
\setlist[description]{nosep, left=0pt, labelsep=0.5em, itemsep=0pt, topsep=0pt, parsep=0pt, partopsep=0pt}

\title{
Tracing LLM Reasoning Processes with Strategic Games: A Framework for Planning, Revision, and Resource-Constrained Decision Making
}

\author{%
  Xiaopeng Yuan\thanks{First author. \texttt{xyuan75@g.ucla.edu}} \\
  \And
  Xingjian Zhang \\
  \And
  Ke Xu \\
  \And
  Yifan Xu \\
  \And
  Lijun Yu \\
  \And
  Jindong Wang \\
  \And
  Yushun Dong \\
  \And
  Haohan Wang\thanks{Corresponding author. \texttt{haohanw@illinois.edu}} \\
}

\begin{document}

\maketitle

\begin{abstract}
% Benchmarks for large language models are falling behind the models they judge. Static question sets can be memorized, and automatically made tests often include mistakes or unpredictable difficulty. Games give us a cleaner test bed. Each game has fixed rules, clear win and loss signals, and instant feedback. This lets us score models without hand grading. We present AdvGameBench, an evaluation platform that drops models into three familiar strategy games: tower defense, auto battler, and turn based combat. Every match runs under the same resource budget and records each move and line of text. This setup lets us track more than the final score. We can watch how a model plans, revises ideas, and respects cost limits. Our metrics look at strategy quality, rate of correction, success of those corrections, and how well the model stays within budget. AdvGameBench is open, reproducible, and simple to extend with new games or rules. By linking automatic feedback with tight budget limits, it turns play into a controlled lab for studying belief revision and other higher level skills. We aim to move the field away from static quizzes and toward dynamic evaluations that are easy to read and rich in feedback.

Large language models (LLMs) are increasingly applied to tasks that require complex reasoning. While most benchmarks focus on evaluating final reasoning outcomes, they overlook the internal processes that lead to those outcomes—such as how a model plans, revises, and makes decisions under constraints. We argue that evaluating these internal reasoning steps is essential for understanding model behavior and improving reliability in real-world applications. To make these processes observable and measurable, we propose using strategic games as a natural and effective environment. These games operate within closed, rule-based systems and provide interpretable states, limited resources, and automatic feedback. 
% This structure makes them especially well-suited for revealing multi-step reasoning dynamics, including strategy formation, revision behavior, and budget sensitivity, without requiring human annotation. 

Therefore, we propose a framework to evaluate LLMs along three core process dimensions: planning, revision, and resource-constrained decision making. 
To support this, we introduce a set of evaluation metrics that extend beyond traditional win rates, incorporating measures such as Over-correction Risk Rate, correction success rate, improvement slope, and over-budget ratio. 

In a set of 4320 adversarial rounds across 12 state-of-the-art models, we find that ChatGPT-o3-mini, which demonstrates strong planning capabilities, achieves the highest composite process score (74.7\%win rate, 78.6\% correction success, and a +0.041 improvement slope). In contrast, Qwen-Plus, despite a high Overcorrection Risk Score of 81.6\%, wins only 25.6\% of its matches, primarily due to excessive resource use. We also observe a negative correlation between Over-correction Risk Rate and correction success rate (Pearson r = –0.51, p = 0.093), suggesting that more frequent corrections do not always improve outcomes. This pattern may reflect impulsive revision strategies, where premature edits reduce overall effectiveness, while more selective approaches lead to greater accuracy. We hope this work offers a new direction for LLM evaluation—focusing not just on what models decide, but on how they decide it.
\end{abstract}

\section{Introduction}

%\hwc{redid with new storyline, please update, and fill in the references. }
Large language models (LLMs) are now capable of solving increasingly complex reasoning tasks \citep{huang2023, zhang2024}. As their performance on traditional benchmarks improves, it has become clear that measuring \textit{outcome accuracy} alone is no longer sufficient. In many real-world scenarios, the quality of an LLM’s reasoning depends not only on the final answer, but also on the internal processes it uses to arrive there: how it plans, how it revises mistakes, and how it makes decisions under resource constraints.

We argue that understanding these \textit{reasoning processes} is a necessary next step in LLM evaluation. Current benchmarks—such as GSM8K \citep{cobbe2021training} or MMLU—offer single-turn questions and measure correctness in isolation. They provide limited visibility into how a model generates hypotheses, updates them in response to feedback, or adjusts its strategy over time. Automatically generated questions have been proposed to avoid memorization \citep{yu2024reeval}, but these bring their own issues, such as variable difficulty and occasional invalidity \citep{minderer2021revisiting, zhang2024careful}.

To address this, we propose shifting the evaluation paradigm: from static, outcome-based tests toward dynamic, process-aware environments. We identify \textbf{strategic games} as a particularly well-suited testbed for this purpose. Games provide closed, rule-based environments with clear feedback signals, bounded resources, and interpretable decision traces. Their structure allows us to directly observe and quantify multi-step reasoning behaviors—without requiring human annotations or handcrafted evaluation rubrics.

In this work, we introduce \textbf{AdvGameBench}, a process-based evaluation framework that embeds LLMs in interactive, resource-constrained strategy games. Rather than judging success solely by win rates, our framework traces how models form plans, revise them when needed, and operate under strict resource budgets. We define a set of core evaluation dimensions—\textbf{planning}, \textbf{revision}, and \textbf{resource-constrained decision making}—and propose concrete metrics that capture each of them.

To support broad and interpretable analysis, AdvGameBench spans three classic game genres—tower defense, auto battler, and turn-based combat—each chosen to expose different cognitive and strategic demands. The framework logs full model outputs and action traces, enabling detailed inspection of decision quality, revision behavior, and adherence to constraints.

\vspace{1em}
\noindent Our key contributions are:
\begin{itemize}
\item A formalization of reasoning process dimensions: planning, revision, and resource-constrained decision making.
\item A game-based evaluation framework that instantiates these dimensions using closed, interpretable, and reproducible environments.
\item A suite of evaluation metrics that measure not only whether models succeed, but how they reason through the task.
\end{itemize}

\section{Related work}
\textbf{LLMs in Gaming Applications.} LLMs have rapidly evolved and demonstrated significant capabilities across various complex tasks, including gaming scenarios\citep{sudhakaran2023}\citep{xu2023werewolf}\citep{gupta2023}\citep{nananukul2024}\citep{light2023avalonbench}. Early studies mainly investigated their performance in text-based adventure games. For example, \citet{tsai2023} examined the capabilities of ChatGPT within interactive fiction. 
% identifying both promising performance and limitations in state tracking and goal-oriented tasks. 
Subsequent research expanded to strategic and multi-agent scenarios. Notably, \citet{akata2023} explored repeated two-player interactions such as the Prisoner’s Dilemma, highlighting the models' strengths in cooperative scenarios and coordination challenges.

Recent attention has increasingly shifted toward multiplayer and complex card games. \citet{yim2024} studied Guandan, a sophisticated Chinese card game characterized by imperfect information and team cooperation. Their research demonstrated that prompting LLMs with Theory of Mind-like strategies significantly improved collaborative performance, but also revealed critical gaps in managing long-horizon states. Similarly, \citet{hu2025} proposed GameArena, a benchmark designed to evaluate fine-grained reasoning skills of LLMs through specialized interactive games.

\textbf{Existing Benchmarks for LLM Evaluation} Despite these advancements, current benchmarks rely predominantly on simplified textual or stylized environments.AppWorld \citep{trivedi2024} , GTBench \citep{duan2024}, GAMEBENCH \citep{costarelli2024}, MINT \citep{wang2023mint}, and AgentBench \citep{liu2023agentbench} illustrate established efforts focusing on puzzle, multi-turn interactions, or agent-oriented tasks. Furthermore, \citet{yang2023starcraft} provided benchmarks specifically for StarCraft II, showcasing sophisticated summarization techniques in strategic gaming contexts. Another research direction evaluates strategic reasoning using game-theoretic frameworks, demonstrating how sophisticated models like GPT-4 \citep{openai2024} approximate human decisions, but often fail to achieve a true rational equilibrium in adversarial or coordination-focused scenarios \citep{lore2023, fan2023}.

Multimodal and embodied approaches have also emerged as significant subfields, exemplified by works such as Voyager in Minecraft \citep{wangvoyager2023} and evaluations of the use of the LLM tool \citep{yang2023}. However, these approaches primarily tackle open-world exploration or general-purpose tasks rather than structured competitive scenarios common in mainstream gaming genres like tower defense or auto battlers. In addition, they often require frequent API interactions or repeated prompts, raising practical cost and latency concerns \citep{wang2024, chiang2024}.

\textbf{In contrast to previous benchmarks} that rely on tool-calling or open-ended exploration, \textbf{AdvGameBench} evaluates LLMs within strategic, rule-based environments where decision-making processes are directly observable. The framework eliminates external dependencies, imposes explicit budget constraints, and embeds models in turn-based adversarial settings. This design enables systematic analysis of not only final outcomes but also intermediate behaviors. 
% such as planning, revision, and adaptation under pressure.

\section{Method}

\subsection{Multi model adversarial structure}

\label{sec:method}
% \begin{figure}[ht]
%   \centering
%   % Extension optional; LaTeX will pick the .pdf
%   \includegraphics[width=\linewidth]{New Diagram -- Xingjian.drawio}
%   \caption{End-to-end \textbf{AdvGameBench} pipeline.  
%   \textbf{1.~Collection \& Game Suites} gather unit, skill, and effect data for three environments (tower-defense, battle-card, turn-based).  
%   \textbf{2.~Model Pool \& Config} lists the 12 LLMs evaluated under a shared decoding setup (temperature 0.3, top-p 1).  
%   \textbf{3.~Adversarial Engine} schedules 4320 matches, allocates roles, runs the simulator, checks budgets, provides feedback, and allows optional revision.  
%   \textbf{4.~Logging} records every trace, result, and resource use in a round database.  
%   Downstream modules compute the five core metrics—OBR, Slope, WR, CSR, CR—and the first-mover analysis (FMA).  
%   Solid arrows denote data flow; dashed arrows show the revision loop triggered by negative feedback.}
%   \label{fig:sample}
% \end{figure}

\begin{wrapfigure}{r}{0.7\textwidth}
    \centering
    \vspace{-12pt}
    \includegraphics[width=\linewidth]{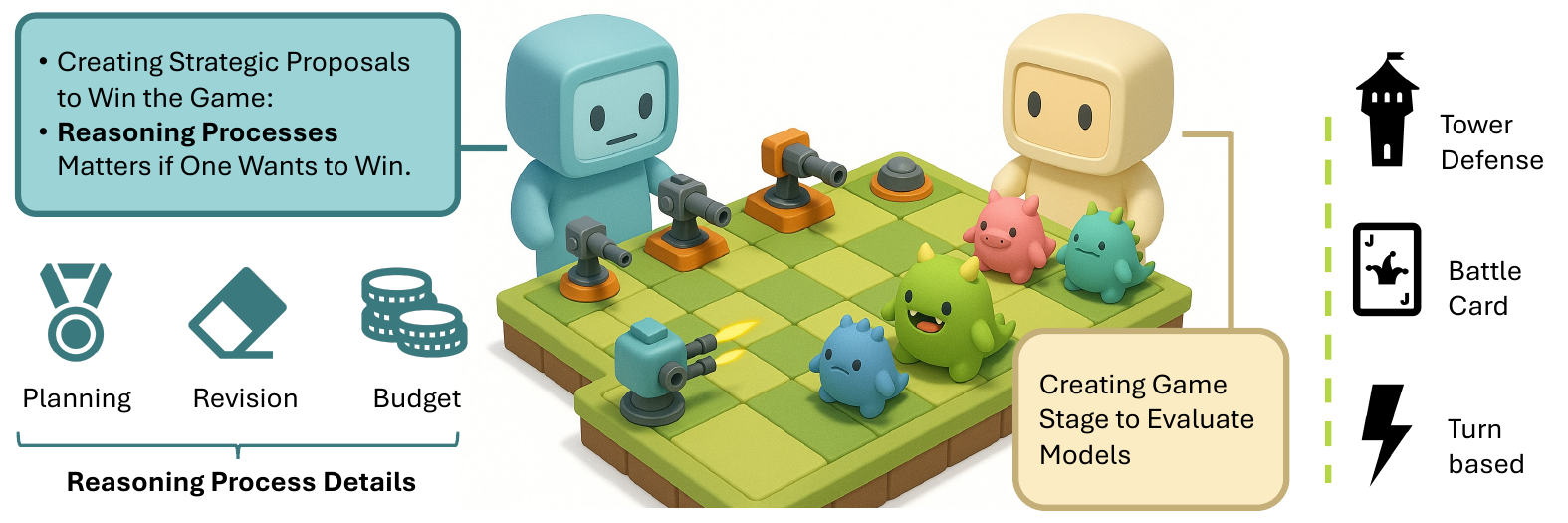}
    \caption{This figure illustrates the AdvGameBench evaluation pipeline. Three strategic game genres—tower defense, auto-battler, and turn-based combat—form the core environments for model evaluation. In each round, the model generates a strategy based on explicit rules; the simulator executes the strategy and returns an outcome; the model optionally revises its plan; and this process iterates over multiple rounds. These interactions yield behavioral trajectories from which process-level metrics are computed, including win rate, over-correction risk rate, correction success rate, and over-budget rate.}
    \vspace{-12pt}
    \label{fig:main}
\end{wrapfigure}

We introduce a structured adversarial framework for evaluating LLMs' \textbf{process-level reasoning behaviors}---specifically, how models \textbf{plan}, \textbf{revision}, and \textbf{resource-constrained decision making} in rule-constrained environments.

\textbf{Game-based, closed-loop setting.} Each evaluation match embeds two LLMs in a \textit{closed, deterministic game simulator} governed by explicit rules and resource constraints. Models receive identical prompts and independently generate strategies. The simulator executes both strategies and returns a rule-verifiable win/loss outcome.

\textbf{Role alternation across diverse games.} We construct three adversarial games—\textit{tower defense, auto-battler, and turn-based combat}—each targeting a distinct reasoning capability. In every round, models alternate between attacker and defender roles, exposing both offensive and defensive strategy formation.

\textbf{Feedback-driven revision.} After each round, models receive outcome-based feedback. They may optionally revise their strategy. These revision sequences are logged and scored using \textbf{process-aware metrics} including correction success rate, over-correction risk, and improvement slope.

\textbf{Control for asymmetry.} To eliminate bias, we evaluate each model pair under both move orders. This ensures symmetry and isolates model-specific behaviors from structural advantages.

\textbf{Adversarial matrix evaluation.} The complete setup yields a dense match matrix, covering all model pairs, roles, and move orders. This enables \textit{systematic comparison} of revision dynamics, constraint adherence, and planning robustness under matched conditions.

\textbf{Original game reimplementations.} All game environments are re-implemented with shifted design from popular games to avoid strategy leakage from popular games. This ensures that models cannot rely on memorized heuristics or latent familiarity with existing game patterns, preserving the objectivity of the evaluation. All these code will be public available.

\subsection{Game suites: Tower Defense, Auto‑battler, Turn‑based}
% We selected three distinct strategic game environments to systematically evaluate the belief revision capabilities of competing LLMs. Each environment represents a popular game genre: tower defense (e.g., Plants vs. Zombies), auto-battler (e.g., Hearthstone Battlegrounds), and turn-based strategy (e.g., Pokémon).
To evaluate how LLMs revise and adapt across varied reasoning contexts, we design three game environments that span distinct forms of strategic complexity. Each environment imposes different constraints and interaction patterns: \textbf{Tower Defense} emphasizes spatial planning under sequential threats, \textbf{Battle Card} requires resource allocation and composition under outcome uncertainty, and \textbf{Turn-based Game} tests decision consistency across multi-step attribute interactions. This diversity ensures that our evaluation covers a broad spectrum of process-level reasoning behaviors.

\textbf{Tower Defense Game.} In this environment, models alternate between attacker and defender roles. Defenders place units on an 11-column battlefield to block attackers advancing from the right. Attackers aim to reach the left boundary, while defenders strive to destroy all attackers. Success criteria and rule violations provide clear feedback for iterative strategy refinement (see \hyperref[app:tower_defense_game]{A.1}).

\textbf{Battle Card Game.} Models control units with distinct attributes: attackers prioritize damage, defenders emphasize protection and recovery. Units engage in automated battles, with combat sequence determined by the number of units each side possesses. The side that eliminates all opposing units first wins, offering explicit outcome-based feedback for model improvement (see \hyperref[app:battle_card_game]{A.2}).

\textbf{Turn-based Attribute Game.} Each side controls three characters with assigned elemental attributes (Fire, Wood, Water, Earth, Light, Dark), featuring strategic interactions based on attribute strengths and weaknesses. Characters choose three skills within a budget constraint, cycling through them in combat. Duels continue until one side remains, clearly indicating the strategic effectiveness and compliance of each model’s choices (see \hyperref[app:turn_based_attribute_game]{A.3}).

\subsection{Evaluation metrics}

To evaluate LLMs beyond raw outcomes, we define a set of metrics tracing how models revise strategies, manage constraints, and adapt over time. 
% This process-aware approach draws from prior studies on dynamic LLM evaluation frameworks such as AdvGameBench \citep{hu2025gamearena, xu2023werewolf, duan2024gtbench}, which emphasize that success in complex tasks depends not only on final outputs but also on the reasoning trajectory that produces them.
We categorize our metrics into three groups:
\begin{itemize}
    \item \textbf{Outcome metric}: measures overall performance.
    \item \textbf{Revision behavior metrics}: assess how models respond to failure.
    \item \textbf{Constraint adherence metrics}: quantify rule compliance under resource limits.
\end{itemize}

\textbf{Win Rate (WR).} 
Win Rate measures the proportion of matches a model wins out of all played games, with rule violations resulting in immediate forfeiture. This metric captures the final outcome of the reasoning process and provides a baseline for comparison. It reflects how well a model integrates planning, revision, and constraint handling into an executable solution.

\textbf{Over-Correction Risk Rate (ORR).}
ORR captures how frequently a model reacts to negative feedback with a revised proposal. This metric targets a critical behavior: over-adjustment in response to failure signals. In practical settings, excessive self-editing can reduce decision stability and degrade coherence over long horizons. High ORR indicates a lack of strategic confidence or an overly reactive revision policy. The need to track this behavior is grounded in the observation that models can degrade their own solutions through unnecessary changes, even when initial plans are viable. 

\textbf{Correction Success Rate (CSR).}
CSR measures how often a revision leads to an improved result—either by eliminating a rule violation or by turning a loss into a win. This metric isolates the effectiveness of the model’s internal feedback loop: can it not only detect failure but also recover from it? A model that frequently edits without reliably improving demonstrates superficial adaptivity rather than meaningful self-correction.

\textbf{Improvement Slope ($\beta$).}
Improvement Slope captures whether a model improves over repeated interactions in matched environments. This measures whether the model can adapt its planning based on prior failures against a fixed opponent type. Unlike static metrics, $\beta$ traces whether a model learns or degrades over time. A flat or negative slope suggests overfitting or myopic adjustment; a positive slope reflects effective cumulative reasoning.

\textbf{Over-Budget Rate (OBR).}
OBR measures how often a model generates proposals that exceed explicit resource constraints. This metric directly evaluates a model’s ability to integrate symbolic or numerical limits into its reasoning process. Many LLMs can optimize performance under unconstrained conditions, but OBR reveals whether they can internalize hard boundaries and behave accordingly. This behavior is essential for real-world deployment, where compliance with external rules is not optional but required for safe execution.

Together, these metrics provide complementary views into different layers of model behavior: WR evaluates final success; ORR and CSR analyze revision dynamics; $\beta$ measures adaptation over time; and OBR enforces structural discipline. Further detailed metrics are discussed in Appendix~\ref{app:additional_metrics}.

\section{Results}
\label{sec:results}
We evaluate 12 leading LLMs, including DeepSeek-R1/V3 \citep{deepseek2024}, Qwen-Plus/Max \citep{bai2023}, Claude-3.5-Sonnet \citep{anthropic2024}, ChatGPT-4.1/4o/o3/o3-mini \citep{openai2024}, Gemini-2/2.5-Flash \citep{anil2025gemini}, and LLaMA-3-70B \citep{grattafiori2024llama}. All models use the same decoding settings: temperature 0.3 and top-$p$ 1, allowing for controlled but non-deterministic generation \citep{renze2024}.

To assess robustness, each model was tested against three diverse opponents: ChatGPT-4o, Claude-3.5-Sonnet, and DeepSeek-V3. This setup avoids evaluation bias caused by shared architectures or training data. In each round, models play against all opponents in turn-based games, with the platform logging win/loss results and correction behaviors for downstream analysis.

\subsection{Revision behavior: correction rate \& success}

\begin{table}[ht]
  \centering
  \footnotesize
  \caption{Benchmark Metrics (WR = win rate, ORR = over-correction risk, CSR = correction success)}
  \label{tab:retention_similarity}
  \label{tab:results}
  \begin{tabular}{l@{\hskip .3em}ccc@{\hskip .3em}ccc@{\hskip .3em}ccc@{\hskip .3em}|@{\hskip .3em}ccc}
    \toprule
  Model
    & \multicolumn{3}{c}{TDG}
    & \multicolumn{3}{c}{BCG}
    & \multicolumn{3}{c}{TAG}
    & \multicolumn{3}{c}{avg}\\
  \cmidrule(lr){2-4} \cmidrule(lr){5-7}\cmidrule(lr){8-10}\cmidrule(lr){11-13}
  & WR & ORR & CSR
  & WR & ORR & CSR
  & WR & ORR & CSR
  & WR & ORR & CSR\\
  \midrule
    ChatGPT-4.1       & 45.0 & 85.7 & 40.0 & 52.5 & 69.7 & 65.2 & 57.5 & 82.4 & 67.9 & 51.7 & 79.4 & 56.8 \\
    ChatGPT-4o        & 65.8 & 81.8 & 55.6 & 60.8 & 44.0 & 63.6 & 59.1 & 82.4 & 46.4 & 58.6 & 70.4 & 52.6 \\
    ChatGPT-o3        & 75.8 & 41.1 & 57.1 & 76.7 & 50.0 & 88.9 & 70.0 & 30.0 & 66.7 & 74.2 & 40.0 & 73.7 \\
    ChatGPT-o3-mini   & 63.3 & 25.9 & 57.1 & 74.2 & 31.6 & 100.0 & 86.7 & 9.0 & 100.0 & 74.7 & 24.5 & 78.6 \\
    Claude-3-5-Sonnet & 56.7 & 89.3 & 56.0 & 45.8 & 70.0 & 64.3 & 55.0 & 76.9 & 65.0 & 52.5 & 77.7 & 61.6 \\
    DeepSeek-R1       & 70.8 & 53.6 & 80.0 & 49.2 & 32.2 & 40.0 & 80.0 & 83.3 & 70.0 & 66.7 & 48.4 & 63.3 \\
    DeepSeek-V3       & 43.3 & 84.6 & 45.5 & 23.3 & 75.5 & 24.3 & 56.7 & 75.0 & 38.1 & 41.1 & 78.4 & 35.2 \\
    Gemini-2-Flash    & 15.8 & 90.6 & 10.4 & 49.2 & 65.7 & 60.9 & 38.3 & 67.5 & 28.0 & 34.4 & 76.8 & 27.1 \\
    Gemini-2.5-Flash  & 60.0 & 40.0 & 60.0 & 59.0 & 79.2 & 68.4 & 58.1 & 76.2 & 56.3 & 62.5 & 65.7 & 63.0 \\
    LLaMA-3-70B       & 33.3 & 90.2 & 29.7 & 42.5 & 76.3 & 51.7 & 65.0 & 69.2 & 66.7 & 46.9 & 80.0 & 45.2 \\
    Qwen-Max          & 39.2 & 44.7 & 5.8 & 10.8 & 50.0 & 10.3 & 41.7 & 51.3 & 36.9 & 30.5 & 48.9 & 16.9 \\
    Qwen-Plus         & 19.2 & 78.4 & 20.0 & 16.7 & 81.5 & 13.6 & 40.8 & 86.1 & 45.2 & 25.6 & 81.6 & 24.3 \\
    \bottomrule
    
  \end{tabular}
\end{table}

Table~\ref{tab:results} shows the win rate, over-correction risk rate, and correction success rate for evaluated models.

\textbf{Win Rate.} ChatGPT-o3-mini and ChatGPT-o3 achieved the highest win rates at 74.7\% and 74.2\%, respectively, substantially outperforming all other models. These results suggest strong capabilities in planning and decision-making under adversarial conditions. In contrast, models such as the Qwen series and Gemini-2-Flash exhibited significantly lower win rates, indicating weaker performance in high-pressure strategic settings.

\textbf{Over-correction Risk Score.} This metric reflects a model’s tendency to overreact to feedback through frequent revisions. Qwen-Plus, DeepSeek-V3, and Claude-3-5-Sonnet exhibited high Over-correction Risk Rates (ORR), suggesting an unstable decision-making process characterized by impulsive or excessive adjustments. In contrast, ChatGPT-o3-mini maintained a relatively low ORR of 49.3\%, indicating a more disciplined and stable strategy that avoids unnecessary revisions unless a confident improvement is identified.

\textbf{Correction Success Rate.} This measures the effectiveness of the attempted revisions. ChatGPT-o3-mini achieved the highest success rate at 78.6\%, indicating that most of its corrections were accurate. Conversely, Qwen-Max and Qwen-Plus had success rates around 20\% despite frequent corrections, reflecting a tendency toward uninformed or premature changes—what we refer to as “blind correction.”

These findings highlight an important distinction: frequent correction behavior does not necessarily imply improved performance. High-performing models engaged in fewer revisions, but these were more targeted and successful. In contrast, models that frequently attempted corrections without sufficient understanding failed to translate effort into meaningful gains. Effective revision thus requires not just responsiveness, but discernment in identifying when and how to intervene.

\subsection{Planning ability: Init‑win \& Improvement Slope}

\begin{figure}[ht]
  \centering
  \begin{minipage}[t]{0.33\textwidth}
    \centering
    \includegraphics[width=\linewidth]{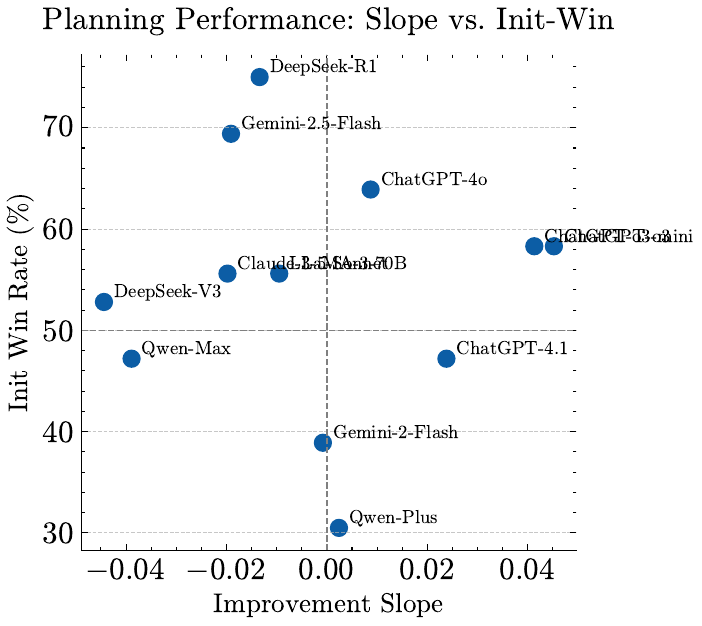}
    \caption{Planning performance: slope vs.\ initial win rate.}
    \label{fig:planning-scatter}
  \end{minipage}%
  \hfill
  \begin{minipage}[t]{0.65\textwidth}
    \centering
    \includegraphics[width=\linewidth]{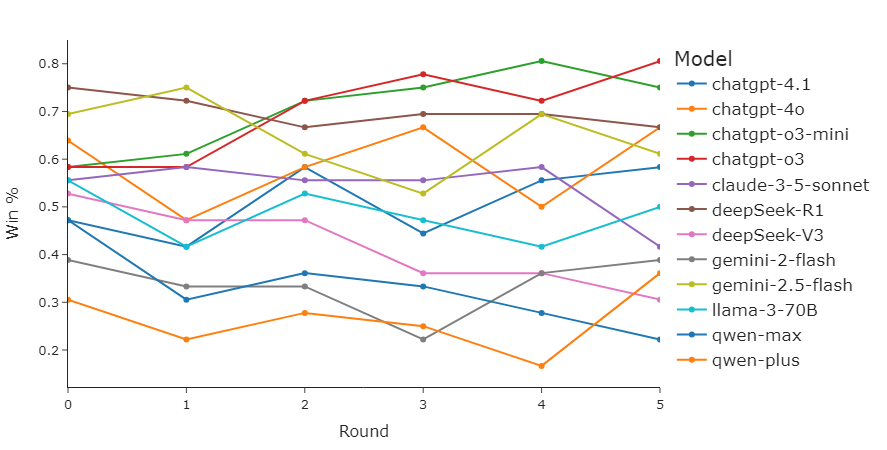}
    \label{fig:output}
    \caption{Win-rate trajectories across five rounds.}
    \label{fig:winrate-curve}
  \end{minipage}
\end{figure}

We evaluate planning capabilities using two complementary metrics: initial win rate (init-win), which reflects first-round performance without feedback, and improvement slope, which measures a model’s ability to enhance its strategy over time. Together, they capture a model’s capacity to start strong and adapt through interaction. \textbf{Figure~\ref{fig:planning-scatter}} shows win rate trajectories across five rounds, while \textbf{Figure~\ref{fig:winrate-curve}} reports improvement slopes.

DeepSeek-R1 achieves the highest init-win (75.0\%) but declines over time, suggesting rigid strategy design. In contrast, ChatGPT-o3 and o3-mini start with lower win rates (58.3\%) yet steadily improve, indicating flexible planning. Models like Gemini-2.5-Flash and Claude-3.5-Sonnet perform well initially but regress, likely due to static heuristics. Qwen models show little progress, pointing to weak feedback integration. Across families, only ChatGPT models consistently improve, reflecting stronger adaptation. These patterns show that robust planning requires not just strong openings, but the ability to refine strategies under pressure—a key dimension captured by process-level metrics like the improvement slope.

\subsection{Resource-constrained decision making}

\begin{wrapfigure}{r}{0.3\textwidth}
  \centering
  \vspace{-10pt}
  \includegraphics[width=\linewidth]{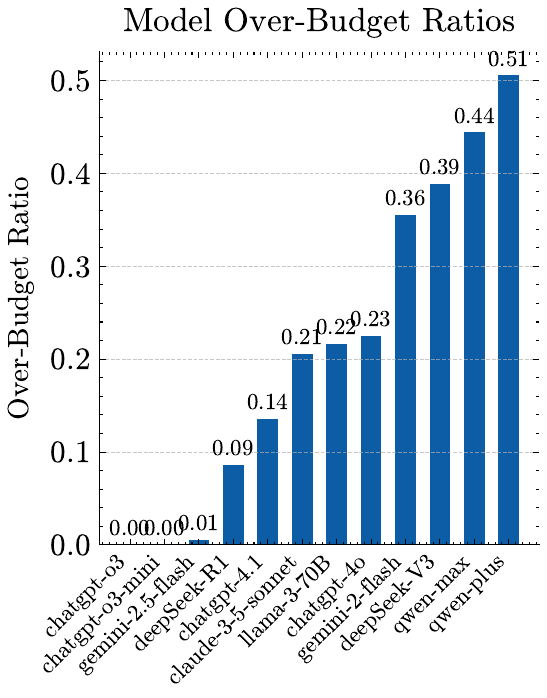}
  \caption{Over-Budget Ratio for Each Model}
  \vspace{-10pt}
  \label{fig:over-budget}
\end{wrapfigure}
\textbf{Figure~\ref{fig:over-budget}} reports the Over-Budget Ratio (OBR), which quantifies the proportion of turns in which a model exceeds the environment's resource constraints. While most models stay within budget in over 80\% of turns, the variation across models is notable. ChatGPT-o3 and ChatGPT-o3-mini maintain perfect budget adherence, never exceeding the allowed limits. In contrast, Qwen-Plus surpasses its budget in approximately half of its turns, and Qwen-Max records similarly high overuse (OBRs of 0.50 and 0.45, respectively). This pattern is strongly aligned with performance: the o3 series models not only exhibit the lowest OBRs but also achieve the highest win rates (74.7\% and 74.2\%), whereas the Qwen models, with the highest OBRs, perform worst in terms of win rate (30.5\% and 25.6\%).

We further find a strong negative correlation between OBR and win rate (Pearson r = –0.95, p < 0.001), indicating that effective resource management is closely tied to model success. High OBRs are often associated with reactive, post-hoc revisions—corrections made after poor initial decisions—which typically fail to compensate for early mistakes. Conversely, models with low OBRs demonstrate more disciplined planning and efficient execution, avoiding costly errors in the first place. These results position OBR as a meaningful process-level indicator that goes beyond outcome accuracy, revealing how well models translate abstract constraints into concrete and consistent decision-making. Strong performers not only remain within budget but also allocate their resources strategically, contributing to higher correction success and overall coherence in behavior.

\subsection{Does revising more really help?}

\begin{figure}[!htb]
  \centering
  \begin{subfigure}{0.24\textwidth}
    \includegraphics[width=\linewidth]{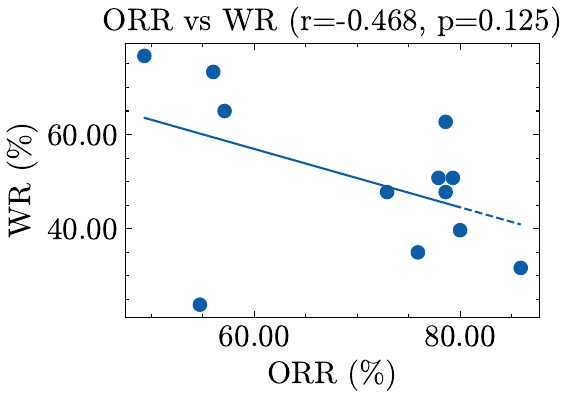}
    \caption{ Correlation between ORR and WR}
    \label{fig:CORR1}
  \end{subfigure}
  \hfill
    \begin{subfigure}{0.24\textwidth}
    \includegraphics[width=\linewidth]{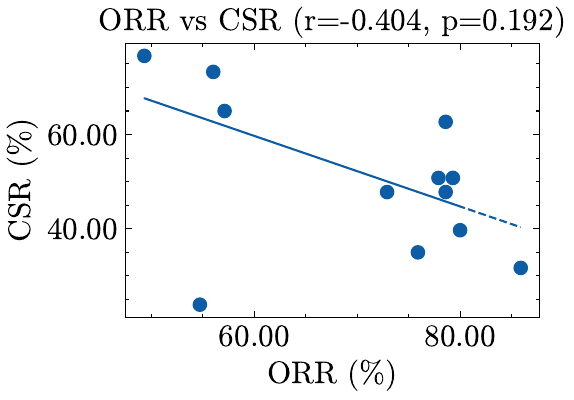}
    \caption{Correlation between ORR and CSR}
    \label{fig:CORR2}
  \end{subfigure}
  \hfill
  \begin{subfigure}{0.24\textwidth}
    \includegraphics[width=\linewidth]{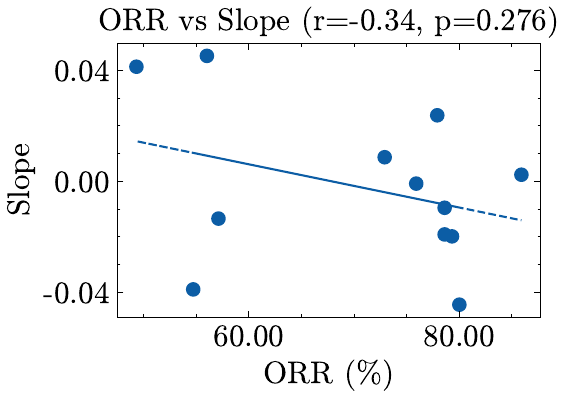}
    \caption{Correlation between ORR and Slope}
    \label{fig:CORR3}
  \end{subfigure}
  \hfill
  \begin{subfigure}{0.24\textwidth}
    \includegraphics[width=\linewidth]{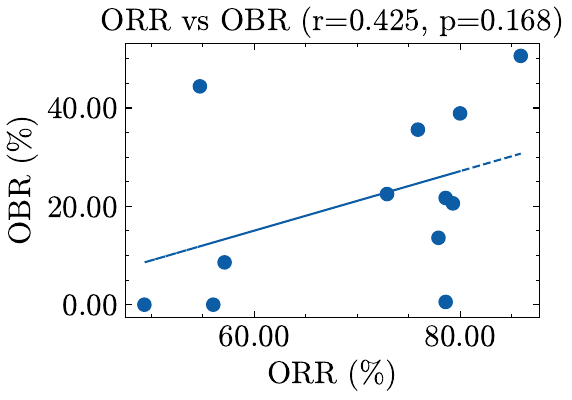}
    \caption{Correlation between ORR and OBR}
    \label{fig:CORR4}
  \end{subfigure}
  \caption{Correlation analysis between over-correction risk rate (ORR) and four main metrics across models}
  \label{fig:CORR_all}
\end{figure}

We quantify a model’s tendency to revise reactively using the \textbf{over‑correction risk rate (ORR)}—the probability that a model submits a new strategy immediately after receiving explicit negative feedback. \textbf{Figure\ref{fig:CORR_all}} presents the correlation between ORR and four process-level outcomes. We observe a moderate negative relationship between ORR and final win rate ($r = -0.47$, $p = 0.13$), suggesting that models which revise more frequently tend to achieve lower overall success. Similarly, ORR correlates negatively with improvement slope ($r = -0.34$, $p = 0.28$), indicating that frequent edits do not accelerate strategic refinement. In terms of budget use, models with higher ORR values are more likely to exceed resource constraints (OBR; $r = +0.43$, $p = 0.17$), and also show lower correction success rates ($r = -0.34$, $p = 0.28$), implying that high-frequency revision may undermine the quality of attempted corrections.

Although none of these effects reach conventional thresholds for statistical significance due to the limited sample size ($n = 12$), the consistency in directional trends is notable. Across all four measures, models with high over-correction risk tend to perform worse: they are less efficient, less successful overall, and less disciplined in their resource usage. In contrast, top-performing models such as \textsc{ChatGPT‑o3‑mini} pair a low ORR with high correction success and zero budget violations. These results highlight a key insight: \textbf{precision in revision—not frequency—is the hallmark of effective strategy adjustment}.

\subsection{Role symmetry and first-move bias}
\begin{figure}[ht]
  \centering
  \begin{subfigure}{0.32\textwidth}
    \includegraphics[width=\linewidth]{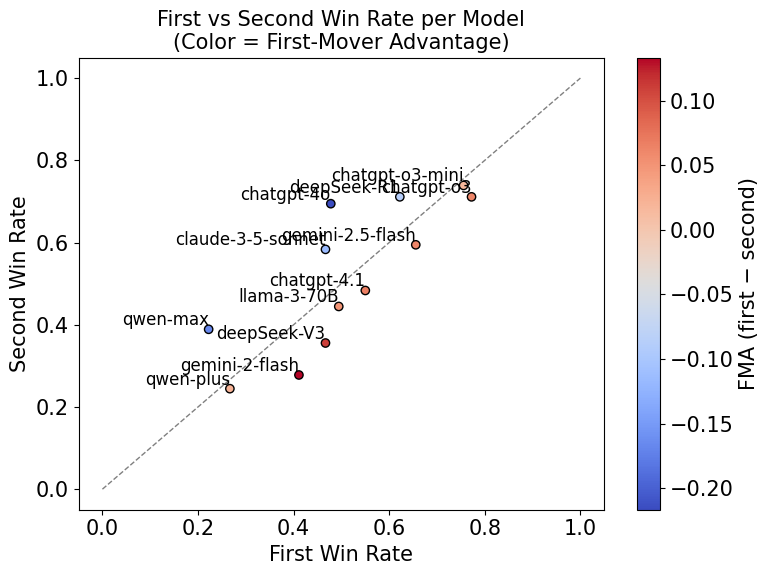}
    \caption{Win rate: first-move vs. second-move}
    \label{fig:FMA1}
  \end{subfigure}
  \hfill
  \begin{subfigure}{0.32\textwidth}
    \includegraphics[width=\linewidth]{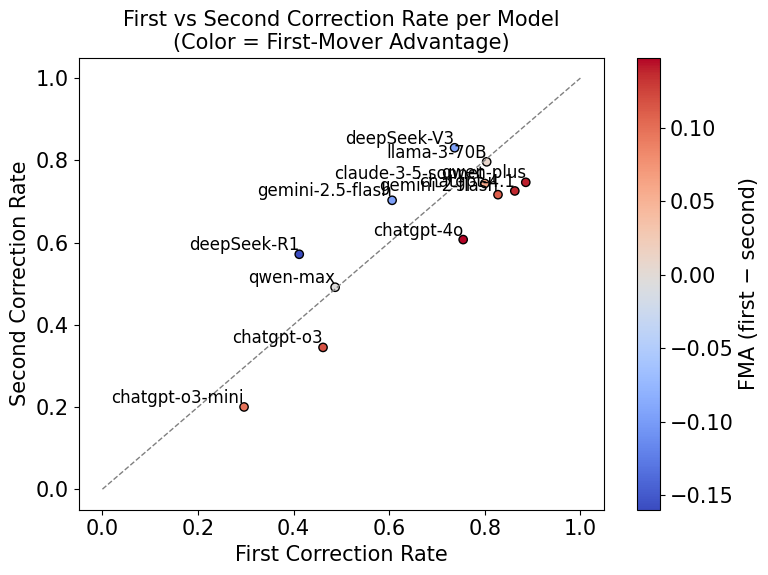}
    \caption{Over-correction risk rate: first-move vs. second-move}
    \label{fig:FMA2}
  \end{subfigure}
  \hfill
  \begin{subfigure}{0.32\textwidth}
    \includegraphics[width=\linewidth]{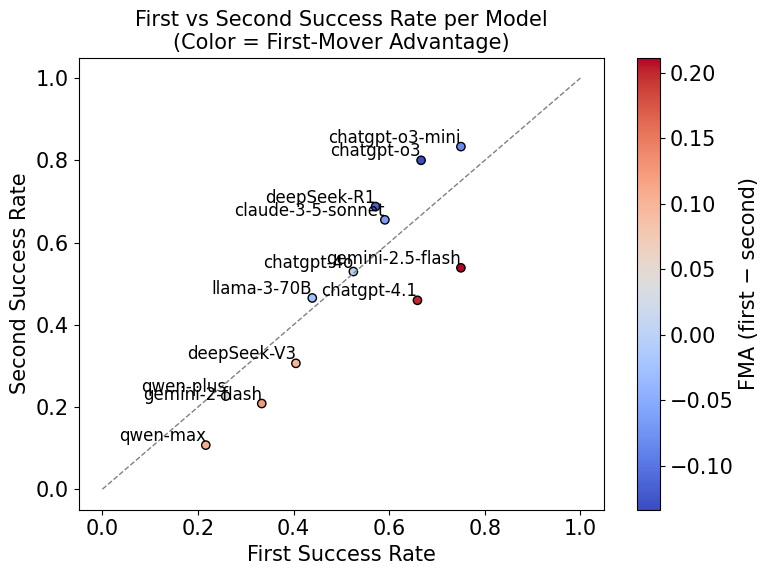}
    \caption{Correction success rate: first-move vs. second-move}
    \label{fig:FMA3}
  \end{subfigure}
  \caption{First-mover advantage (FMA) across three behavioral dimensions.}
  \label{fig:FMA_all}
\end{figure}
We use First-Mover Advantage (FMA) to examine how model performance differs when initiating an action versus responding to a prior move. We analyze this effect across three dimensions: win rate, over-correction risk rate, and correction success rate. As shown in Figure~\ref{fig:FMA1}, most models exhibit relatively minor differences in win rate between first- and second-mover roles, with FMA values generally within five percentage points. This suggests limited systematic advantage based on turn order for overall success. However, several models deviate from this trend. Gemini-2-Flash (FMA = +13.2\%) performs substantially better when acting first, while ChatGPT-4o (FMA = –21.7\%) and Qwen-Max (FMA = –16.7\%) exhibit the opposite pattern, achieving higher win rates when playing second. These results suggest that certain models are more sensitive to the structural asymmetries introduced by move order.

Stronger patterns emerge when examining correction behavior. In Figure~\ref{fig:FMA2} and Figure~\ref{fig:FMA3}, we observe that most models show a clear preference for initiating rather than responding. For example, ChatGPT-4o and ChatGPT-4.1 demonstrate significantly higher over-correction risk rates when acting first (FMA = +14.8\% and +13.8\%, respectively). Similarly, first-mover performance gains are evident in correction success rates for Gemini-2.5-Flash (+21.2\%), Gemini-2-Flash (+12.5\%), and ChatGPT-4.1 (+19.9\%). These findings underscore the importance of accounting for role asymmetry in evaluation setups. Our dual-first configuration helps mitigate first-mover bias, offering a more balanced and interpretable view of model behavior under asymmetric game dynamics.

\subsection{Holistic comparison via radar chart}
\begin{wrapfigure}{r}{0.5\textwidth}
  \centering
  \vspace{-15pt}
  \includegraphics[width=\linewidth]{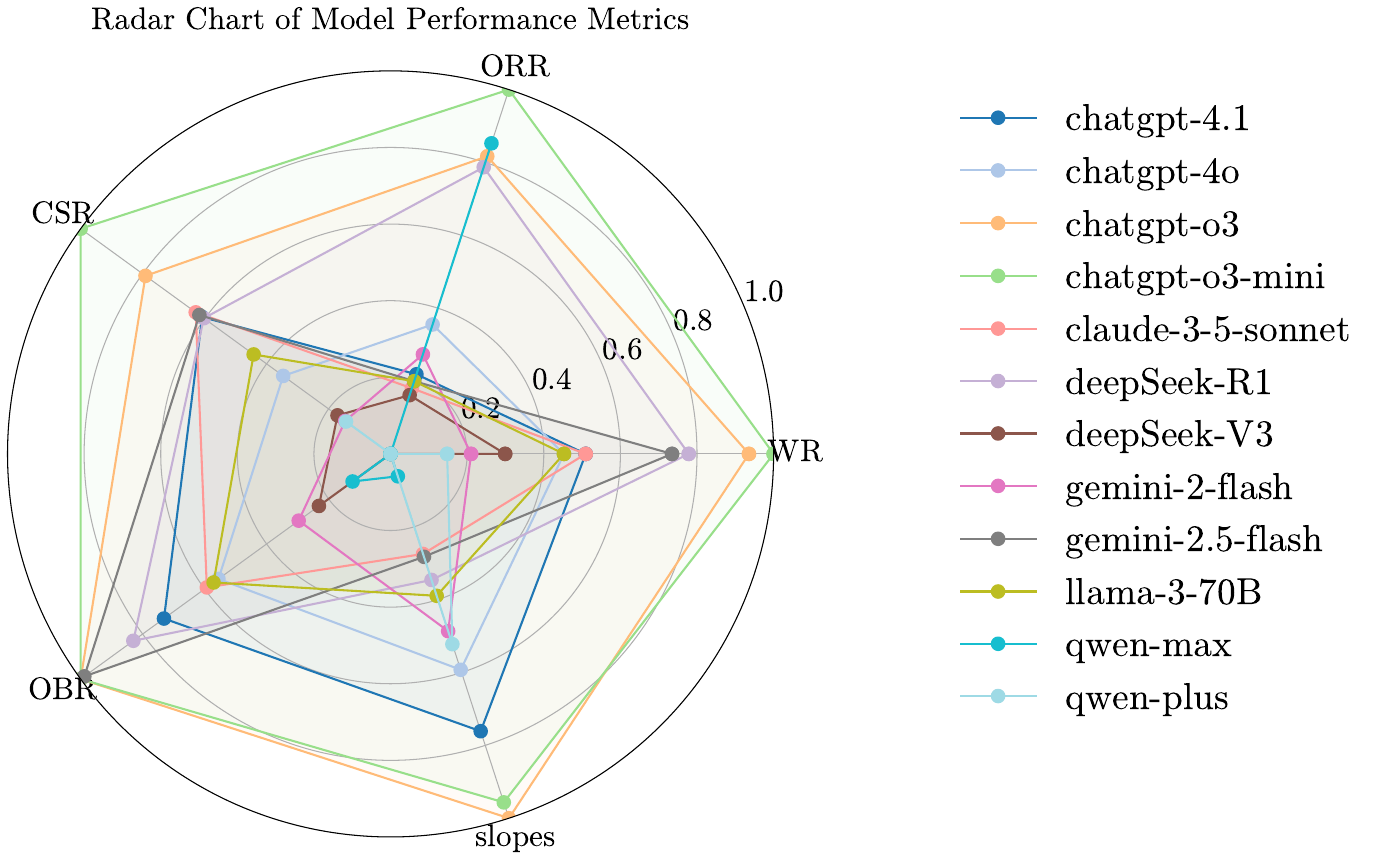}
  \caption{Model performance metrics}
  \vspace{-10pt}
  \label{fig:radar_chart}
\end{wrapfigure}
To synthesize model performance across reasoning dimensions, we constructed a radar chart visualizing five normalized metrics: win rate (WR), correction success rate (CSR), improvement slope, 1 – over-correction risk rate (ORR), and 1 – over-budget rate (OBR). All metrics were scaled to a common range, with inversions applied where necessary so that higher values consistently indicate better performance. This unified view enables a comparative assessment of both outcome and process quality across models.

ChatGPT-o3 and o3-mini form the largest radar areas, reflecting strong, consistent performance across all dimensions. They pair high win rates with effective corrections, stable improvement, and disciplined resource use, indicating well-integrated reasoning. In contrast, models like Qwen-Plus and Qwen-Max show sharp imbalances, marked by frequent but ineffective revisions and frequent budget violations. Gemini models perform moderately in CSR and planning but are similarly constrained by high correction risk or poor budget control. These patterns highlight that top performance requires balance across planning, revision, and constraint adherence—not just isolated strength. Larger radar areas correspond to more robust reasoning pipelines, reinforcing that process quality is essential to understanding model competence.

\subsection{Model-Specific Strengths and Underlying Mechanisms}
Our evaluation shows that different LLMs exhibit distinct process-level strengths, often reflecting differences in architecture and alignment. Models from the ChatGPT family—especially o3 and o3-mini—achieve consistently high win rates while maintaining disciplined correction behavior and strict budget adherence. This pattern suggests a stable internal revision mechanism, likely shaped by reinforcement learning with human feedback (RLHF) and conservative fine-tuning objectives that prioritize reliability over exploration. In contrast, models such as Qwen-Plus and DeepSeek-V3 revise frequently but achieve low correction success and often exceed resource limits. These behaviors point to reactive decision-making and overly eager feedback incorporation, which can destabilize planning over time. We refer to this pattern as “over-correction,” where excessive responsiveness leads to fragmented strategies and reduced overall performance.

Other models, including Claude-3.5-Sonnet and Gemini-2.5-Flash, show more balanced profiles across metrics. While they do not dominate any single dimension, they perform moderately well in planning, correction, and resource management. This may reflect broader instruction-tuning or multitask training that encourages general adaptability without specializing in any one skill. Taken together, these differences underscore that LLM capabilities are shaped by design trade-offs: between caution and flexibility, local reactivity and global coherence. Our process-level metrics—particularly ORR and improvement slope—help reveal these trade-offs, offering a more nuanced view of model behavior than outcome-based evaluations alone. They also provide actionable insights into how alignment strategies and decoding preferences influence long-horizon reasoning, suggesting concrete directions for model development and benchmarking.

\section{Discussion}

\label{sec:discussion}

\vspace{-0.25em}
\noindent
\textbf{Process‐rather‐than‐outcome evaluation.}\;
AdvGameBench purposefully shifts the evaluation lens from \emph{what} an LLM answers to \emph{how} it arrives there.
By embedding models in three rule-bound strategy games, we can observe---and score---their behaviour along the three dimensions defined in Method section (see \hyperref[sec:method]{Method}).:
\emph{planning} (initial strategy quality and improvement slope),
\emph{revision} (correction rate and success),
and \emph{resource-constrained decision making} (budget adherence).

\textbf{Empirical studies.}
Our study of twelve production-scale LLMs across \textbf{4\,752} adversarial rounds yields three consistent findings:
\begin{enumerate}[leftmargin=*,itemsep=2pt]
    \item \textbf{Integrated skill trumps single metrics.}
          Models that balance the three dimensions---notably \textsc{ChatGPT--O3-Mini} with a \num{74.7}\,\% win rate, \num{78.6}\,\% correction-success rate, and positive improvement slope of $+0.0413$---outperform models that excel in only one aspect.
    \item \textbf{“Spray-and-pray” revision is counter-productive.}
          \textsc{Qwen-Plus} issues corrections in \num{81.6}\,\% of error states yet wins only \num{25.6}\,\% of games and overspends in nearly half the turns.
          Across all systems, correction frequency and efficacy are negatively correlated (Pearson $r=-0.51$, $p=0.093$), indicating that \emph{calibrated} self-editing matters more than sheer persistence.
    \item \textbf{Budget fidelity is a leading indicator of success.}
          The two models that never violated resource limits (\textsc{ChatGPT--O3} and \textsc{ChatGPT--O3-Mini}) also posted the highest win rates,
          whereas both Qwen variants combine the largest over-budget ratios with the poorest outcomes.
\end{enumerate}

\textbf{Hallucation.}
In our tower defense experiments, all defensive units were consistently referred to as \textit{soldiers}. However, several models frequently generated the term \textit{peashooter}, which was never introduced in the task instructions. A review of the interaction logs reveals that this phenomenon does not stem from a reasoning failure, but rather from prior associations learned during pretraining—specifically, the frequent co-occurrence of “tower defense” and the game \textit{Plants vs. Zombies} in web-scale corpora. This leads models to default to familiar terminology, even when it conflicts with the defined rules of the environment. Such behavior undermines the validity of the benchmark, effectively turning the evaluation into a test of memorized correlations rather than genuine planning or constraint adaptation. To eliminate this form of memory bias, we redesigned the game environment to neutralize lexical cues and ensure that performance reflects models’ ability to engage with novel rules and dynamic constraints, rather than recalling pretraining artifacts.

\textbf{Limitations.}
\label{para:limitations}
AdvGameBench currently (i) covers three turn-based genres but no real-time or cooperative play,
(ii) logs unit-level actions yet does not attribute win contributions to individual decisions,
and (iii) relies on synthetic opponents, which---although diversified---cannot fully mirror human play styles.
These choices were deliberate to keep the study controlled and reproducible, but they constrain external validity.

% \paragraph{Future work.}
% We plan to (a) extend the benchmark with real-time and team-based games, (b) add per-action credit assignment metrics, and
% (c) investigate curriculum setups where game difficulty adapts online to a model’s competence.
% More broadly, we hope AdvGameBench will stimulate research on training objectives that \emph{explicitly} reward budget fidelity,
% revision calibration, and long-horizon planning.

\section{Conclusion}
\label{sec:conclusion}

\noindent
% Static accuracy benchmarks have become an insufficient proxy for real-world robustness.  
% Deployment-ready systems must also \emph{plan soundly}, \emph{revise judiciously}, and \emph{respect resource constraints}.  
% AdvGameBench meets this need by turning strategic gameplay into an open, extensible laboratory in which those process-level traits can be quantified.

% By exposing the entire decision trace---from initial plan through budgeted actions and self-corrections---the benchmark reveals failure modes that outcome-only tests conceal.  
% These fine-grained signals can guide both diagnostic studies and the design of training objectives that reward disciplined reasoning.  
% Ultimately, we see AdvGameBench as one step toward a broader shift in LLM evaluation: away from asking only \emph{“Did the model answer correctly?”} and toward asking \emph{“How did the model reason, adapt, and stay within bounds while answering?”}  
% Such process-aware scrutiny is essential for building reliable, accountable, and trustworthy language models.

Static accuracy benchmarks have become an insufficient proxy for real-world robustness.
Deployment-ready systems must also \emph{plan soundly}, \emph{revise judiciously}, and \emph{respect resource constraints} to function effectively in practical environments.
AdvGameBench meets this need by turning strategic gameplay into an open, extensible laboratory in which those process-level traits can be systematically quantified, monitored, and improved over time.

By exposing the entire decision trace—from initial plan through budgeted actions and self-corrections—the benchmark reveals failure modes that outcome-only tests often conceal.
These fine-grained signals enable not only diagnostic analysis of model behavior but also principled design of training objectives that reward disciplined, context-sensitive reasoning under pressure.
They also help evaluate whether models can maintain stability across repeated trials, even in adversarial or resource-limited conditions.
AdvGameBench supports controlled ablations, adversarial setups, and resource perturbations, making it a flexible platform for probing model resilience.

Ultimately, we see AdvGameBench as one step toward a broader shift in LLM evaluation: away from asking only \emph{“Did the model answer correctly?”} and toward asking \emph{“How did the model reason, adapt, and stay within bounds while answering?”}
Such process-aware scrutiny is essential for building language models that are not only accurate but also reliable, accountable, and aligned with real-world deployment demands.

\bibliography{main}

\appendix
\section{Appendix}

\subsection{Tower defense game}\label{app:tower_defense_game}

\subsubsection{Game rules}
\begin{enumerate}
    \item Players can purchase characters and place them on the battlefield. The battlefield consists of 5 rows (corresponding to y-coordinates 0-4). The human side can place units in a designated area spanning 11 columns (corresponding to x-coordinates 0-10).
    \item Demons spawn from the right side of the battlefield (x-coordinates 11) and move left. Human units are placed on the left side of the battlefield, remain stationary, and attack approaching enemies.
    \item All units attack according to their attack interval, automatically attacking when their cooldown ends. Defending units fire bullets or activate skills to attack enemies. Invading units engage in melee attacks when they come into contact with defending units.
    \item Each grid cell can only contain one human unit at a time. Placing a new unit in an occupied cell is not allowed.
    \item When an attack hits, the target takes damage based on the attacker's power. If a unit's health drops to 0, it is eliminated and removed from the battlefield.
    \item If all enemies are eliminated, the player wins. If any enemy successfully reaches the left side of the battlefield, the player loses.
\end{enumerate}

\subsubsection{Human units}
\begin{longtable}{p{0.25\textwidth}p{0.75\textwidth}}
\toprule
\textbf{Unit} & \textbf{Attributes} \\
\midrule
HandgunSoldier & Health: 3, Shooting interval: 1000ms, Cost: 100, Damage per shot: 1, No special abilities. \\
\midrule
RifleSoldier & Health: 3, Shooting interval: 500ms, Cost: 200, Damage per shot: 1, No special abilities. \\
\midrule
MachineGunSoldier & Health: 3, Shooting interval: 250ms, Cost: 400, Damage per shot: 1, No special abilities. \\
\midrule
ShieldSoldier & Health: 15, Cost: 50, Only for defense, no attack capabilities. \\
\midrule
EnhancedShieldSoldier & Health: 30, Cost: 100, Only for defense, no attack capabilities and Bouncing Demon cannot jump over. \\
\midrule
FlamethrowerSoldier & Health: 2, Cost: 200, Shooting interval: 1000ms, Damage per shot: 1, Deals an additional 1 damage every 1000ms. \\
\midrule
IceSoldier & Health: 2, Shooting interval: 1000ms, Cost: 200, Damage per shot: 1, Reduces enemy speed by half. \\
\midrule
AntiAirSoldier & Health: 2, Shooting interval: 1000ms, Cost: 175, Damage per shot: 1, Can attack airborne units. \\
\midrule
Bomb & Health: 50, Detonation time: 500ms, Cost: 200, Explosion range: 3×3, Damage per explosion: 30, Destroyed after detonation. \\
\midrule
LinearExplosion & Health: 50, Detonation time: 500ms, Cost: 200, Explosion range: the entire row, Damage per explosion: 30, Destroyed after detonation. \\
\midrule
MagneticSoldier & Health: 2, Shooting interval: 2000ms, Cost: 100, Damage per shot: 0, Releases a magnetic pulse that disables the defensive abilities of ShieldDemon and MachineDemon. \\
\midrule
LightMage & Health: 2, Damage per shot: 0, Cost: 150, No attack capabilities, Changes the attributes of bullets in the same row, converting their damage type to light. \\
\midrule
RocketLauncherSoldier & Health: 2, Shooting interval: 1000ms, Damage per shot: 2, Cost: 600, Launches rockets, dealing damage to enemies within one grid. \\
\bottomrule
\end{longtable}

\subsubsection{Demon units}
Note: A speed of 2 requires 14 seconds to travel from spawn to the last human grid.

\begin{longtable}{p{0.25\textwidth}p{0.75\textwidth}}
\toprule
\textbf{Unit} & \textbf{Attributes} \\
\midrule
NormalDemon & Health: 10, Speed: 2, Attack interval: 1000ms, Cost: 100, Damage per attack: 1, No special abilities. \\
\midrule
GreatDemon & Health: 20, Speed: 2, Attack interval: 1000ms, Cost: 175, Damage per attack: 1, Higher health. \\
\midrule
DemonKing & Health: 100, Speed: 2, Attack interval: 1000ms, Cost: 800, Damage per attack: 5. \\
\midrule
SpeedyDemon & Health: 10, Speed: 4, Attack interval: 1000ms, Cost: 150, Damage per attack: 1, Moves faster. \\
\midrule
ShieldDemon & Health: 10, Speed: 2, Attack interval: 1000ms, Cost: 175, Damage per attack: 1, Takes 70\% less damage from normal attacks. \\
\midrule
MachineDemon & Health: 20, Speed: 2 (increases to 3 when activated), Attack interval: 1000ms, Cost: 250, Damage per attack: 3, Reduced damage due to mechanical body. \\
\midrule
BouncingDemon & Health: 10, Speed: 2, Attack interval: 1000ms, Cost: 150, Damage per attack: 1, Can jump over certain units except for the EnhancedShieldSoldier. \\
\midrule
ShieldBreakerDemon & Health: 10, Speed: 2, Attack interval: 1000ms, Cost: 150, Damage per attack: 1 (×5 against shield units). \\
\midrule
FireDemon & Health: 10, Speed: 2, Attack interval: 1000ms, Cost: 150, Damage per attack: 1, Immune to fire damage. \\
\midrule
FrostDemon & Health: 10, Speed: 2, Attack interval: 1000ms, Cost: 150, Damage per attack: 1, Immune to ice damage and unaffected by slow effects. \\
\midrule
FlyingDemon & Health: 10, Speed: 2, Attack interval: 1000ms, Cost: 200, Damage per attack: 1, Only affected by anti-air attacks and can pass through human units directly. \\
\midrule
ShadowDemon & Health: 10, Speed: 2, Attack interval: 1000ms, Cost: 300, Damage per attack: 1, Can cast dark magic, making same-row allies immune to non-light damage. \\
\midrule
SummoningDemon & Health: 10, Speed: 1, Attack interval: 1000ms, Cost: 300, Damage per attack: 1, Summons a Normal Demon to the left grid every 5000ms. \\
\bottomrule
\end{longtable}

\subsection{Battle card game}\label{app:battle_card_game}

\subsubsection{Game rules}
\begin{enumerate}
    \item At the start of the game, players can purchase all desired characters at once, up to a maximum of 7 characters. Gold characters cost three times as much as bronze characters, but their stats (attack, health, numerical skill effects, etc.) are twice as high. Non-numerical skills are not affected by this multiplier.
    \item Initiative Determination: The side with more characters attacks first. If both sides have the same number of characters, the invader attacks first.
    \item Elemental Advantage: Certain elements have an advantage over others, granting a bonus in combat (Fire $>$ Nature, Nature $>$ Water, Water $>$ Earth, Earth $>$ Fire).
    \item Battle Process: Both sides will attack based on their respective target\_priority (target priority). However, if there are Taunt minions on the opponent's side, attackers must prioritize attacking them. The attack order follows a left-to-right sequence. The first minion in the invaders or defenders list (as defined in the JSON file) will attack first, depending on which side has the initiative. After that, the first minion from the opposing side attacks. Then, the second minion from the attacking side follows, then the second minion from the opposing side, and so on in an alternating pattern. If a minion's health reaches zero, it is eliminated. The battle continues with both sides attacking in turns until one side is completely wiped out, resulting in victory for the other side.
    \item If all characters on one side are eliminated, the other side wins.
    \item If both sides are eliminated simultaneously in the same attack resolution, the Invader wins.
\end{enumerate}

\subsubsection{Invader units}
\begin{longtable}{p{0.25\textwidth}p{0.75\textwidth}}
\toprule
\textbf{Unit} & \textbf{Attributes} \\
\midrule
FireLizard & Attack: 2, Health: 2, Cost: 1, Ability: Deals 2 damage to the enemy that killed it upon death. \\
\midrule
WaterElemental & Attack: 2, Health: 2, Cost: 1, Ability: Gains +1 Attack when attacking. \\
\midrule
PoisonFrog & Attack: 1, Health: 1, Cost: 2, Ability: Instantly destroys any minion it damages. \\
\midrule
MoltenHound & Attack: 3, Health: 1, Cost: 2, Ability: Deals 1 damage to all enemies upon death. \\
\midrule
BattleFrenzy & Attack: 7, Health: 4, Cost: 2, Ability: Each attack reduces its Attack by 4. \\
\midrule
BanditLeader & Attack: 8, Health: 3, Cost: 3, Ability: Any excess damage from an attack carries over to the next target. \\
\midrule
LavaGolem & Attack: 1, Health: 8, Cost: 3, Ability: Forces enemies to attack this minion first, Burns the attacker for 3 damage per turn when hit. \\
\midrule
TideGuardian & Attack: 4, Health: 2, Cost: 3, Ability: Absorbs the first source of damage taken (divine shield), Attacks twice each turn. \\
\midrule
TideLord & Attack: 4, Health: 9, Cost: 5, Ability: Doubles its Attack when taking damage. \\
\midrule
Phoenix & Attack: 5, Health: 5, Cost: 5, Ability: Deals damage equal to its Attack to the target and its adjacent enemies, Revives with full Health after being defeated once per game. \\
\midrule
ShadowOverlord & Attack: 4, Health: 4, Cost: 5, Ability: Summons a Slow Skeleton (3/1) upon death. \\
\bottomrule
\end{longtable}

\subsubsection{Defender units}
\begin{longtable}{p{0.25\textwidth}p{0.75\textwidth}}
\toprule
\textbf{Unit} & \textbf{Attributes} \\
\midrule
Sapling & Attack: 2, Health: 2, Cost: 1, Ability: Gains +1 Health when attacking. \\
\midrule
RockBeetle & Attack: 1, Health: 5, Cost: 1, Ability: Forces enemies to attack this minion before others. \\
\midrule
ForestSeer & Attack: 2, Health: 2, Cost: 2, Ability: At the start of the game, grants +1 Attack and +2 Health to all Nature Allies. \\
\midrule
StoneWarrior & Attack: 2, Health: 5, Cost: 2, Ability: Forces enemies to attack this minion before others. Summons a RockBeetle upon death. \\
\midrule
EliteSoldier & Attack: 1, Health: 1, Cost: 2, Ability: At the start of the game, grants Divine Shield to adjacent minions and +1 Attack. \\
\midrule
Paladin & Attack: 3, Health: 6, Cost: 3, Ability: Has Divine Shield; gains +2 Attack whenever a friendly minion loses its Divine Shield. \\
\midrule
BlackRock & Attack: 5, Health: 1, Cost: 3, Ability: At the start of the game, gains +3 Health for each friendly minion. \\
\midrule
VineProtector & Attack: 5, Health: 4, Cost: 3, Ability: Upon death, restores 2 Health to all friendly minions. \\
\midrule
King & Attack: 3, Health: 10, Cost: 5, Ability: Summons a 2/2 Soldier with Divine Shield whenever it attacks (if there is an open space). \\
\midrule
MountainGiant & Attack: 4, Health: 9, Cost: 5, Ability: Forces enemies to attack this minion first, Reduces the attack of the attacker by 2 when hit. \\
\midrule
AncientTreant & Attack: 4, Health: 4, Cost: 5, Ability: At the start of the game, grants +3 Attack and +3 Health to all allied minions. \\
\bottomrule
\end{longtable}
\subsection{Turn-based attribute game}
\label{app:turn_based_attribute_game}
\subsubsection{Game rules}
\begin{enumerate}
    \item This game is a turn-based character battle game divided into two factions: Invader and Defender. Each faction consists of three characters. The Invader faction includes Fire, Water, and Dark elements, while the Defender faction includes Wood, Earth, and Light elements. Characters appear and act in the order they are listed in the data.
    \item Combat proceeds in rounds. In each round, the three Invader characters act first in order, followed by the three Defender characters. The sequence then repeats in the next round.
    \item Each character has three skills that are used in a preset, looping sequence. On each turn, a character uses the next skill in their list and continues cycling through them in order.
    \item The game features an elemental effectiveness system: Fire beats Wood, Wood beats Earth, Earth beats Water, and Water beats Fire (1.2$\times$ damage when effective, 0.8$\times$ when resisted). Light and Dark counter each other with 1.5$\times$ damage. All other combinations deal the standard 1.0$\times$ damage.
    \item If all characters on one side are eliminated, the other side wins.
\end{enumerate}

\subsubsection{Invader skills}
\begin{longtable}{p{0.25\textwidth}p{0.75\textwidth}}
\toprule
\textbf{Skill Name} & \textbf{Description} \\
\midrule
\multicolumn{2}{l}{\textbf{Fire Skills}} \\
\midrule
flame\_splash & Deals 12 damage and applies Burning for 2 rounds (1 layer, 5 damage per round). Cost: 1 \\
residual\_warmth & Increases the damage of the next fire-based skill by 30\% for 1 round. Cost: 1 \\
burst\_flame\_bomb & Deals 25 base damage, plus 3 additional damage for each Burning layer on the target. Cost: 2 \\
flame\_whirlwind & Applies 4 layers of Burning to the target, lasting 2 rounds. Each layer deals 5 damage per round. Cost: 2 \\
magma\_eruption & Deals 40 base damage, plus 5 extra damage per Burning layer. Removes all Burning after the attack. Cost: 3 \\
hell\_curtain & Deals 35 damage and grants a shield that reflects 30 melee damage, lasting 2 rounds (1 layer). Cost: 3 \\
\midrule
\multicolumn{2}{l}{\textbf{Water Skills}} \\
\midrule
stream\_pierce & Deals 10 damage and grants 1 permanent layer of Tidal Surge. Cost: 1 \\
water\_barrier & Grants a 5-point shield for 3 rounds and increases Tidal Surge by 1 layer. Cost: 1 \\
whirlpool\_strangle & Deals 20 base damage, plus 4 additional damage per Tidal Surge layer. Cost: 2 \\
ice\_branded & Deals 15 damage and causes the target to take 50\% more damage next turn (1 round). Cost: 2 \\
tsunami\_ending & Deals 30 base damage, plus 5 additional damage per Tidal Surge layer. Removes all Tidal Surge after the attack. Cost: 3 \\
abyss\_resonance & Deals 3 damage per Tidal Surge layer and grants a shield worth 6 per layer, lasting 3 rounds. Cost: 3 \\
\midrule
\multicolumn{2}{l}{\textbf{Dark Skills}} \\
\midrule
shadow\_claw & Deals 14 damage and heals the user for 30\% of the damage dealt (rounded down). Cost: 1 \\
fear\_whisper & Reduces the target's damage taken by 10\% for 3 rounds (1 layer). Cost: 1 \\
soul\_siphon & Deals 25 damage. If the target's HP is below 50\%, deals an extra 15 damage. Cost: 2 \\
night\_ambush & Deals 20 damage and causes the target to take 20\% more damage next turn (1 round). Cost: 2 \\
final\_announcment & Deals 45 base damage, plus 5 extra damage for every 10\% HP the target has lost. Cost: 3 \\
void\_assimilation & Sacrifices 20\% of current HP to deal penetration damage equal to twice the HP sacrificed. Cost: 3 \\
\bottomrule
\end{longtable}

\subsubsection{Defender skills}
\begin{longtable}{p{0.25\textwidth}p{0.75\textwidth}}
\toprule
\textbf{Skill Name} & \textbf{Description} \\
\midrule
\multicolumn{2}{l}{\textbf{Wood Skills}} \\
\midrule
bud\_healing & Grants Bud Healing for 3 rounds, restoring 6 HP per round. Cost: 1 \\
parasitic\_seed & Applies Parasitic Seed for 3 rounds, immediately deals 10 damage. The target takes 5 counter damage each time they attack. Cost: 1 \\
life\_totem & Restores 25 HP and grants Life Totem for 3 rounds, increasing healing received by 10\%. Cost: 2 \\
natural\_purification & Removes negative statuses from the user and deals 30 damage to the target. Cost: 2 \\
forest\_reincarnation & Restores 60 HP. If it exceeds max HP, the excess is converted into a shield (50\% of excess HP) for 3 rounds. Also deals 20 damage to an enemy. Cost: 3 \\
poison\_vine & Applies Poison Vine for 3 rounds, dealing 25 damage per round. Cost: 3 \\
\midrule
\multicolumn{2}{l}{\textbf{Earth Skills}} \\
\midrule
rock\_armor & Grants 12 shield for 3 rounds and reflects 5 melee damage while the shield is active. Cost: 1 \\
earth\_shock & Deals 20 damage. Cost: 1 \\
granite\_barrier & Grants Granite Barrier for 3 rounds, decreasing damage by 40\%. Cost: 2 \\
quicksand\_trap & Applies Quicksand Trap for 3 rounds. The target's next 3 damage are delayed by 20\% and each trigger deals 10 damage. Cost: 2 \\
earth\_pulse & Grants shield based on HP lost (8 shield per 10\% HP lost), lasting permanently. Cost: 3 \\
core\_rebound & Deals 80\% of stored damage to the target. Clears stored damage after use. Cost: 3 \\
\midrule
\multicolumn{2}{l}{\textbf{Light Skills}} \\
\midrule
holy\_glimmer & Removes a negative status (if any) and restores 8 HP to the user. Also deals 8 light damage to an enemy. Cost: 1 \\
faith\_emblem & Grants Faith Emblem for 1 round. The next damage taken is reduced by 20\% and converted into healing. When triggered, deals 10 damage to the attacker. Cost: 1 \\
divine\_link & Grants Divine Link for 1 round. The next damage taken is reflected back to the attacker. Cost: 2 \\
luminous\_dispel & Removes one buff from the target (if any) and applies a debuff for 2 rounds that reduces their attack by 15\%. Cost: 2 \\
angelic\_sanctuary & Grants Angelic Sanctuary for 3 rounds, reducing all incoming damage by 30 points. Cost: 3 \\
divine\_sword & Deals 20 damage and grants a buff that increases the next skill's damage by 20. Cost: 3 \\
\bottomrule
\end{longtable}

\section{Additional evaluation metrics}
\label{app:additional_metrics}
This section details supplementary metrics used to provide a more granular understanding of LLM behavior in strategic game environments, complementing the core metrics presented in Section 3.4.

\subsection{Rule violation Rate (RVR)}
This metric measures how often a model's initial strategy proposal fails to adhere to the game's explicit rules, particularly budget constraints.
Let $M_i$ denote model $i$. Let $T_i$ be the total number of initial strategy proposals made by model $M_i$ across all games and rounds where it provides an initial strategy. For each initial strategy proposal $S_{i,t}^{(0)}$ (where $t$ indexes these proposals, $t \in \{1, \dots, T_i\}$), let $V(S_{i,t}^{(0)})$ be an indicator function, such that $V(S_{i,t}^{(0)}) = 1$ if the strategy $S_{i,t}^{(0)}$ violates any game rule (including budget constraints), and $V(S_{i,t}^{(0)}) = 0$ otherwise.
The Rule Violation Rate for model $M_i$ is:
\begin{equation}
\text{RVR}_i = \frac{\sum_{t=1}^{T_i} V(S_{i,t}^{(0)})}{T_i}
\label{eq:app_rule_violation_rate}
\end{equation}
A lower RVR indicates better adherence to explicit constraints during initial planning phases.

\subsection{Constructive Rate (CnstrR)}
This metric assesses whether a correction attempt, following negative feedback, leads to an objectively improved game state, even if it doesn't immediately result in a win or full rule compliance.
Let $E_{i,g,k}$ be the event that negative feedback is received for strategy $S_{i,g,k}$ (model $i$, game instance $g$, $k$-th strategy in that game instance). Let $A_{i,g,k+1}$ be the event that model $M_i$ proposes a new strategy $S_{i,g,k+1}$ in response. Let $\Phi(S)$ be a game-specific state evaluation function where higher values indicate a more advantageous position for the model (e.g., based on remaining unit health/cost difference, reduced enemy threat, or other heuristic measures of game state quality). A correction $S_{i,g,k+1}$ is considered constructive if $\Phi(S_{i,g,k+1}) > \Phi(S_{i,g,k})$.
The Constructive Rate for model $M_i$ is:
\begin{equation}
\text{CnstrR}_i = \frac{\sum_{g=1}^{G_i} \sum_{k=0}^{K_{i,g}-1} \mathbb{I}\left(E_{i,g,k} \land A_{i,g,k+1} \land \left(\Phi(S_{i,g,k+1}) > \Phi(S_{i,g,k})\right)\right)}{\sum_{g=1}^{G_i} \sum_{k=0}^{K_{i,g}-1} \mathbb{I}\left(E_{i,g,k} \land A_{i,g,k+1}\right) + \varepsilon}
\label{eq:app_constructive_rate}
\end{equation}
where $G_i$ is the total number of game instances involving model $M_i$ where corrections are possible, $K_{i,g}$ is the number of strategies proposed by model $M_i$ in game instance $g$, $\mathbb{I}(\cdot)$ is the indicator function, and $\varepsilon$ is a small constant to prevent division by zero. This captures the tendency for revisions to make incremental, positive progress. Calculating $\Phi(S)$ requires domain-specific heuristics tailored to each game environment.

\subsection{Multi-aspect Strategic Similarity Ratio (\texorpdfstring{$\mathcal{S}_{\text{MASR}}$}{S	extsubscript{MASR}})}
This metric assesses the similarity between a corrected strategy $S^{(k+1)}$ and the preceding strategy $S^{(k)}$ by considering multiple facets: structural, semantic, and functional similarity. For a given model $M_i$, let $S_{i,g,k}$ be the $k$-th strategy in game instance $g$.
Let $\text{Sim}_{\text{struct}}(S', S)$, $\text{Sim}_{\text{sem}}(S', S)$, and $\text{Sim}_{\text{func}}(S', S)$ be normalized similarity scores in $[0, 1]$ for these aspects:
\begin{itemize}
    \item \textbf{Structural Similarity ($\text{Sim}_{\text{struct}}$)}: Measures overlap in concrete elements (e.g., unit types, positions, configurations). This can be quantified using metrics like the Jaccard index on sets of chosen units/actions, or a normalized graph edit distance if strategies are represented as graphs $G(S)$. For instance, $\text{Sim}_{\text{struct}}(S', S) = 1 - \frac{\text{GED}(G(S'), G(S))}{\text{max\_GED}}$, where GED is graph edit distance.
    \item \textbf{Semantic Similarity ($\text{Sim}_{\text{sem}}$)}: Measures similarity in the underlying strategic intent or concept, often derived from embeddings of textual descriptions or structured representations of the strategy. If $\mathbf{e}(S)$ is an embedding vector for strategy $S$, then $\text{Sim}_{\text{sem}}(S', S) = \max(0, \text{cosine\_similarity}(\mathbf{e}(S'), \mathbf{e}(S)))$.
    \item \textbf{Functional Similarity ($\text{Sim}_{\text{func}}$)}: Measures overlap in the intended strategic functions or roles fulfilled by the strategy components (e.g., defensive formations, offensive pushes, resource gathering focus). If $\mathcal{F}(S)$ is the set of strategic functions embodied by strategy $S$, then $\text{Sim}_{\text{func}}(S', S) = \frac{|\mathcal{F}(S') \cap \mathcal{F}(S)|}{|\mathcal{F}(S') \cup \mathcal{F}(S)| + \epsilon'}$, where $\epsilon'$ prevents division by zero for empty sets.
\end{itemize}
The multi-aspect similarity ratio for a specific correction from $S_{i,g,k}$ to $S_{i,g,k+1}$ is a weighted combination:
\begin{equation}
\mathcal{S}_{\text{MASR}}(S_{i,g,k+1}, S_{i,g,k}) = \sum_{j=1}^{N_{\text{aspects}}} \theta_j \cdot \text{Sim}_{\text{aspect}_j}(S_{i,g,k+1}, S_{i,g,k})
\label{eq:app_multi_aspect_similarity_instance}
\end{equation}
where $N_{\text{aspects}}$ is the number of similarity aspects (e.g., 3 for structural, semantic, functional), and $\theta_j$ are weights such that $\sum_{j=1}^{N_{\text{aspects}}} \theta_j = 1$ and $\theta_j \ge 0$.
The average $\bar{\mathcal{S}}_{\text{MASR}}(i)$ for model $M_i$ is calculated over all valid correction steps:
\begin{equation}
\bar{\mathcal{S}}_{\text{MASR}}(i) = \frac{\sum_{g=1}^{G_i} \sum_{k=0}^{K_{i,g}-1} \mathbb{I}(E_{i,g,k} \land A_{i,g,k+1}) \cdot \mathcal{S}_{\text{MASR}}(S_{i,g,k+1}, S_{i,g,k})}{\sum_{g=1}^{G_i} \sum_{k=0}^{K_{i,g}-1} \mathbb{I}(E_{i,g,k} \land A_{i,g,k+1}) + \varepsilon}
\label{eq:app_multi_aspect_similarity_avg}
\end{equation}
This metric quantifies the degree of strategic preservation or alteration during revisions. A high $\bar{\mathcal{S}}_{\text{MASR}}(i)$ indicates a tendency towards conservative revision, while a low value suggests more aggressive or radical strategy changes.

\subsection{First-Mover Advantage (FMA)}
First-Mover Advantage (FMA) quantifies the performance difference for a model when it acts first (initiates the interaction or round) versus when it acts second (responds to the opponent's initial move). This can be calculated for various performance metrics $X$, such as Win Rate (WR), Over-correction Risk Rate (ORR), or Correction Success Rate (CSR).
Let $M_i$ be the model under evaluation. Let $\mathcal{G}_{i,\text{first}}$ be the set of game instances where model $M_i$ moved first, and $\mathcal{G}_{i,\text{second}}$ be the set of game instances where model $M_i$ moved second. Let $N_{i,\text{first}}(X) = |\mathcal{G}_{i,\text{first}}|$ and $N_{i,\text{second}}(X) = |\mathcal{G}_{i,\text{second}}|$ be the respective counts of such game instances for which metric $X$ is applicable.
Let $X_{i,m}$ be the value of metric $X$ observed for model $M_i$ in a specific game instance $m$.
The average performance for model $M_i$ on metric $X$ when moving first is:
\begin{equation}
\bar{X}_{i, \text{first}} = \frac{1}{N_{i,\text{first}}(X) + \varepsilon} \sum_{m \in \mathcal{G}_{i,\text{first}}} X_{i,m}
\label{eq:app_fma_first}
\end{equation}
Similarly, the average performance for model $M_i$ on metric $X$ when moving second is:
\begin{equation}
\bar{X}_{i, \text{second}} = \frac{1}{N_{i,\text{second}}(X) + \varepsilon} \sum_{m \in \mathcal{G}_{i,\text{second}}} X_{i,m}
\label{eq:app_fma_second}
\end{equation}
where $\varepsilon$ is a small positive constant to prevent division by zero if a model never plays in one of the roles or if the metric is not applicable in those instances.
The First-Mover Advantage for metric $X$ and model $M_i$ is then defined as the difference:
\begin{equation}
\text{FMA}_X(i) = \bar{X}_{i, \text{first}} - \bar{X}_{i, \text{second}}
\label{eq:app_fma_difference}
\end{equation}
A positive $\text{FMA}_X(i)$ indicates that model $M_i$ performs better on metric $X$ when it has the first-move advantage. Conversely, a negative value suggests better performance when moving second. The magnitude of $\text{FMA}_X(i)$ indicates the strength of this turn-order bias.

\bibliographystyle{apalike}

\end{document}